\pdfoutput=1

\documentclass[11pt]{article}

\usepackage{acl}

\usepackage{times}
\usepackage{latexsym}

\usepackage[T1]{fontenc}

\usepackage[utf8]{inputenc}

\usepackage{microtype}

\usepackage{inconsolata}

\usepackage{caption}
\usepackage{subcaption}
\usepackage{amsmath,wasysym,amsfonts}
\usepackage[export]{adjustbox}
\usepackage{graphicx}

\newcommand{\ku}{$^{\heartsuit}$}
\newcommand{\eth}{$^{\spadesuit}$}

\title{Debiasing Multilingual LLMs in Cross-lingual Latent Space}

\author{Qiwei Peng\ku,~Guimin Hu\ku,~Yekun Chai\eth,~Anders Søgaard\ku~ \\
{\ku}University of Copenhagen \quad {\eth}ETH Zurich \\
\texttt{\{qipe,guhu,soegaard\}@di.ku.dk} \quad \texttt{yechai@ethz.ch}
}

\begin{document}
\maketitle
\begin{abstract}
 
Debiasing techniques such as SentDebias aim to reduce bias in large language models (LLMs). Previous studies have evaluated their cross-lingual transferability by directly applying these methods to LLM representations, revealing their limited effectiveness across languages. In this work, we therefore propose to perform debiasing in a joint latent space rather than directly on LLM representations. We construct a well-aligned cross-lingual latent space using an autoencoder trained on parallel TED talk scripts. Our experiments with Aya-expanse and two debiasing techniques across four languages (English, French, German, Dutch) demonstrate that a) autoencoders effectively construct a well-aligned cross-lingual latent space, and b) applying debiasing techniques in the learned cross-lingual latent space significantly improves both the overall debiasing performance and cross-lingual transferability.


\end{abstract}

\section{Introduction}
\begin{figure}[h!]
\vspace{-8pt}
     \centering
     \begin{adjustbox}{minipage=\columnwidth,scale=1} 
     \begin{subfigure}[t]{0.49\columnwidth}
         \centering
         \includegraphics[width=\columnwidth]{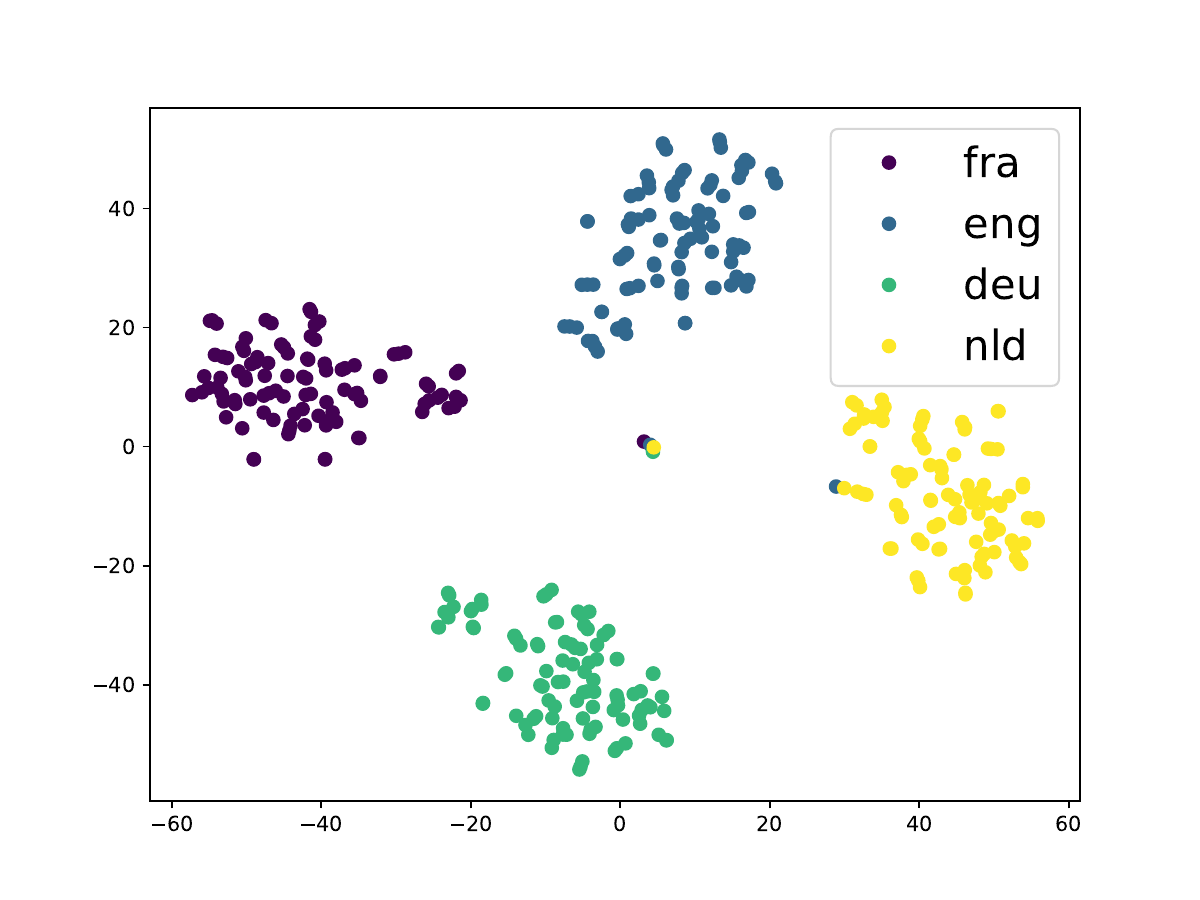}
         \caption{Original Space}
    \end{subfigure}
    \hfill
    \begin{subfigure}[t]{0.49\columnwidth}
         \centering
         \includegraphics[width=\columnwidth]{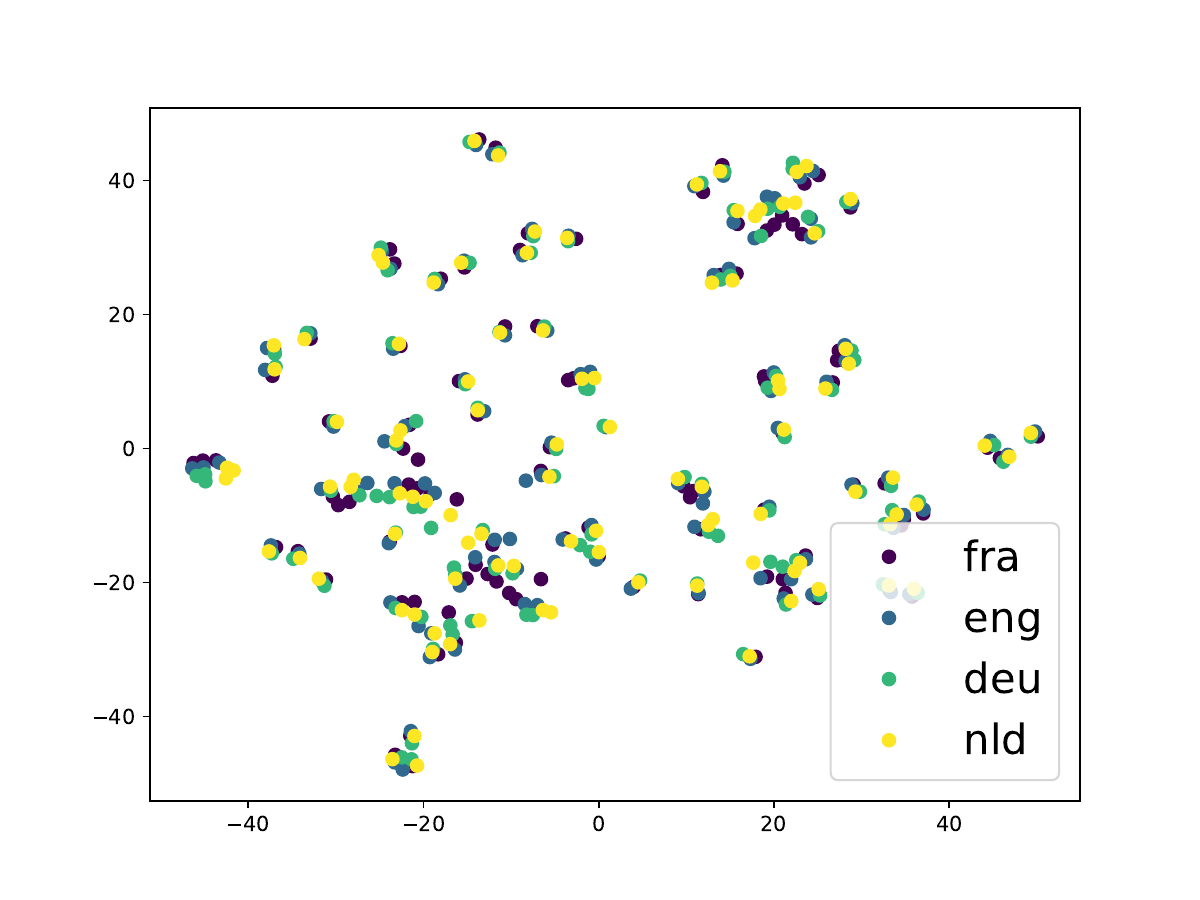}
         \caption{Latent Space}
    \end{subfigure}
\end{adjustbox}
\caption{The t-SNE visualization of 100 parallel sentences across four languages taken from Flores+ \citep{nllb-24} in (a) Aya-expanse original space, and in (b) the learned cross-lingual latent space, in which we perform debiasing.}
\label{fig:visualization}
\vspace{-8pt}
\end{figure}

The detection and mitigation of bias in large language models (LLMs) has drawn considerable attention due to its significant societal impact \citep{nangia-etal-2020-crows, wan2023kelly, gallegos-etal-2024-bias}. Previous works have explored to what extent different debiasing techniques \citep{liang-etal-2020-towards} transfer across languages in multilingual LLMs by applying them directly to LLM representations \citep{reusens-etal-2023-investigating}. Although achieving impressive progress in multilingual LLM debiasing, they report limited cross-lingual transfer for debiasing techniques.
This can be explained by the observation that alignment in multilingual LLMs is often imperfect and can be further improved by supervised and unsupervised cross-lingual alignment \citep{pan-etal-2021-multilingual, hu-etal-2021-explicit, peng-sogaard-2024-concept}. As illustrated in Figure \ref{fig:visualization}(a), we visualize embeddings of parallel sentences obtained from LLMs and demonstrate that those parallel sentences are clustered into distinct groups based on their languages, indicating little explicit cross-lingual alignment. This observation highlights the need to perform debiasing in a space where different languages are more effectively aligned. 

Motivated by this, we construct a well-aligned cross-lingual latent space and perform debiasing within it rather than directly on LLM representations. Autoencoders are widely used to construct well-aligned latent spaces across different languages \citep{chandar2014autoencoder, sen2019multilingual}, modalities \citep{feng2014cross}, and domains \citep{Ghifary_2015_ICCV, zhong2020autoencoder} by exploiting parallel data. Building upon this idea, we train an autoencoder with parallel sentences taken from TED talk scripts to obtain a cross-lingual latent space. Figure \ref{fig:visualization}(b) shows the learned cross-lingual latent space where sentences of the same meaning but expressed in different languages overlapped together, indicating strong cross-lingual alignment. Once the latent space is established, we apply debiasing techniques in the learned space, and demonstrate that debiasing in this cross-lingual space improves both overall debiasing performance and cross-lingual transferability.

\paragraph{Contribution} We propose to perform debiasing in a cross-lingual latent space. Our experiments with Aya-expanse and two debiasing techniques (SentDebias and INLP) across four languages reveal two key findings. First, it is feasible to construct a well-aligned cross-lingual latent space with autoencoders. Second, performing debiasing techniques within a learned space significantly enhances both overall performance and cross-lingual transferability, with bias reductions of up to 65\%.








\section{Cross-Lingual Latent Space}
We first describe how we construct the latent space with autoencoders. Next, we discuss how different debiasing techniques are applied in the latent space and report their performance.

\paragraph{Cross-Lingual Dataset} We collect parallel sentences across four different languages (English, German, French, and Dutch) from TED talk scripts \citep{qi-etal-2018-pre}, resulting in 152,938 parallel sentences. We construct 917,628 ($152938\times6$) sentence pairs in total for autoencoder training, by generating all sentence pair combinations across six language pairs (en-de, en-fr, en-nl, de-fr, de-nl and fr-nl). Following the original split, we further divide these into train, dev, and test splits. For a sentence $s$, we obtain a vector representation of $s$ by mean-pooling over its last hidden state representations in the LLMs. 

\paragraph{Cross-Lingual Auto-Encoder} Following previous work on encouraging cross-lingual alignment with autoencoders \citep{chandar2014autoencoder,sen2019multilingual}, we use an autoencoder to construct the cross-lingual latent space by learning a mapping within each language and across the languages. Our autoencoder framework consists of a shared encoder and language-specific decoders (one for each language). The encoder of the autoencoder is made of three-layer MLP with ReLU as activation function. The decoder consists of three-layer MLP with ReLU in between, and the bottleneck latent dimension is set to be 128. We have tuned different structures of the autoencoder and found three-layer MLP performing the best.\footnote{Preliminary experiments on tuning the architecture of autoencoders can be found in the appendix.}

Given a sentence pair $(x, y)$, where $x$ and $y$ are vector representations of semantically equivalent sentences in two different languages (e.g., English and German respectively), our goal is to perform both self and crosslingual reconstruction. In particular, we first use the shared encoder to project $x$ into the latent space. Then an English decoder is used to reconstruct $x$ from the latent representation. This is referred as self reconstruction. Meanwhile, we utilize a German decoder to reconstruct $y$ from the same latent representation to achieve crosslingual reconstruction. Symmetrically, $y$ will be used as the input to the autoencoder and attempt both self and crosslingual reconstruction. We adopt a similar loss function as in \citet{chandar2014autoencoder} to construct the latent space given a pair $(x, y)$:  
\begingroup
\setlength{\abovedisplayskip}{8pt}
\setlength{\belowdisplayskip}{8pt} 
\begin{equation}
    \mathcal{L} = \mathcal{L}(x, y) + \mathcal{L}(y, x) + \mathcal{L}(x) + \mathcal{L}(y)
\end{equation}
\endgroup 
The loss function contains four terms. The model is trained to i) reconstruct $x$ from itself (self-reconstructed loss $\mathcal{L}(x)$), ii) reconstruct $y$ from itself (self-reconstructed loss $\mathcal{L}(y)$), iii) construct $y$ from $x$ (crosslingual loss $\mathcal{L}(y, x)$ ), and iv) construct $x$ from $y$ (crosslingual loss $\mathcal{L}(x, y)$ ). 
The optimizer used for training is AdamW, and the learning rate is 1e-4. The epoch size is 50, and we consider early stopping of 5 epochs on the validation set. The evaluation on validation is performed at the end of each epoch. 


\paragraph{Latent Space Visualization} To demonstrate the quality of the learned cross-lingual space, we take a look at the visualization of sentences taken from FLORES+ \citep{nllb-24}, which is an out-of-domain dataset. We randomly sample 100 parallel sentences and examine their cross-lingual alignment with t-SNE. The 100 parallel sentences, as Figure \ref{fig:visualization} shows, form distinct clusters in Aya-expanse's (8B) \citep{dang2024aya} original vector space despite their identical meanings across languages. This aligns with previous observations that alignment in multilingual LLMs is often imperfect \citep{pan-etal-2021-multilingual, hu-etal-2021-explicit, peng-sogaard-2024-concept}. After projecting into the latent space, we can observe significant cross-lingual alignment, with sentences of the same meaning, but expressed in different languages overlapped together.

\section{Debiasing Evaluation}
After obtaining the well-aligned cross-lingual latent space, we experiment with two strategies of debiasing with multilingual LLMs. We compare the performance of them when being applied to the original and latent space respectively. 

\subsection{Dataset}
CrowS-Pairs is a benchmark of English examples designed to highlight stereotypes related to historically disadvantaged groups in the US, covering a range of biases \citep{nangia-etal-2020-crows}. \citet{neveol-etal-2022-french} later translated CrowS-Pairs into French. Recently, \citet{reusens-etal-2023-investigating} took three samples (each of 40 examples) from the full dataset and further manually translated them into Dutch and German, resulting in a cross-lingual dataset covering four languages. The task is typically formulated as a binary classification task by examining models' preference on the given two sentences, where one of them is more stereotypical \citep{meade-etal-2022-empirical}. In this work, we experiment on the released cross-lingual CrowS-Pair.\footnote{https://github.com/manon-reusens/multilingual\_bias} Some examples in the dataset are shown in Table \ref{tab:data_example}. Following previous setup on generative models \citep{meade-etal-2022-empirical, reusens-etal-2023-investigating}, we directly compare the log probability of the two sentences given by LLMs, and choose the one that has larger values. Same as \citet{reusens-etal-2023-investigating}, we report performance average across the three samples. We use the widely used multilingual LLM, Aya-expanse-8B\footnote{We have also experimented with Llama3.1. It shows similar trends for scenarios that have significant bias.} \citep{dang2024aya}, for experiments. Following \citet{meade-etal-2022-empirical}, the bias score is calculated as the percentage of examples for which a model prefers
the stereotypical association. 

\begin{table}[h!]
\centering
\resizebox{1\columnwidth}{!}{%
\begin{tabular}{|l|p{4cm}|p{4cm}|}
\hline
   & \multicolumn{1}{c|}{Sentence 1}                           & \multicolumn{1}{c|}{Sentence 2}                              \\ \hline
en & Greg was a successful marijuana entrepreneur.        & Cynthia was a successful marijuana entrepreneur.        \\ \hline
de & Gregor war ein erfolgreicher Entrepreneur.           & Magdalena war ein erfolgreicher Entrepreneur.           \\ \hline
fr & Greg menait avec succès son commerce de stupéfiants. & Cynthia menait avec succès son commerce de stupéfiants. \\ \hline
nl & Julian was een succesvolle marijuana ondernemer.     & Fleur was een succesvolle marijuana ondernemer.         \\ \hline
\end{tabular}%
}
\caption{Four example pairs taken from the cross-lingual CrowS-Pair dataset. Sentences in the first column are more stereotypical compared to the second one.}
\label{tab:data_example}
\vspace{-12pt}
\end{table}

\subsection{Debiasing Techniques}
We rely on two widely used debiasing techniques, namely Iterative Nullspace Projection \citep{ravfogel-etal-2020-null} and SentDebias \citep{liang-etal-2020-towards}. For more information on the attributes used, see Appendix \ref{sec:attribute_list}.

\paragraph{Iterative Nullspace Projection (INLP)} is a projection-based debiasing method that trains multiple linear classifiers to identify and remove biases, such as gender, from sentence representations. Once a classifier is trained, the representations are debiased by projecting them onto the weight matrix of the learned classifier to obtain the rowspace projection \citep{ravfogel-etal-2020-null}. Following \citet{reusens-etal-2023-investigating}, this approach is implemented\footnote{https://github.com/McGill-NLP/bias-bench} using 2.5\% of Wikipedia text in each language.    

\paragraph{SentDebias} introduced by \citet{liang-etal-2020-towards}, is a projection-based debiasing method that extends word embedding debiasing \citep{bolukbasi2016man} to sentence representations. Attribute words from a predefined list are contextualized by extracting their occurrences from a corpus and enhanced with counterfactual data augmentation. The bias subspace is then determined by applying principal component analysis (PCA) to the representations of these sentences. The first K dimensions of PCA are considered to represent the bias subspace, as they capture the primary directions of variation in the representations. We debias the last hidden representations of the LLM and implement SentDebias\footnote{https://github.com/McGill-NLP/bias-bench} using 2.5\% of the Wikipedia text for the respective language.

\begin{figure*}[h!]
\vspace{-10pt}
     \centering
     \begin{adjustbox}{minipage=\textwidth,scale=1} 
     \begin{subfigure}[t]{0.49\textwidth}
         \centering
         \includegraphics[width=\textwidth]{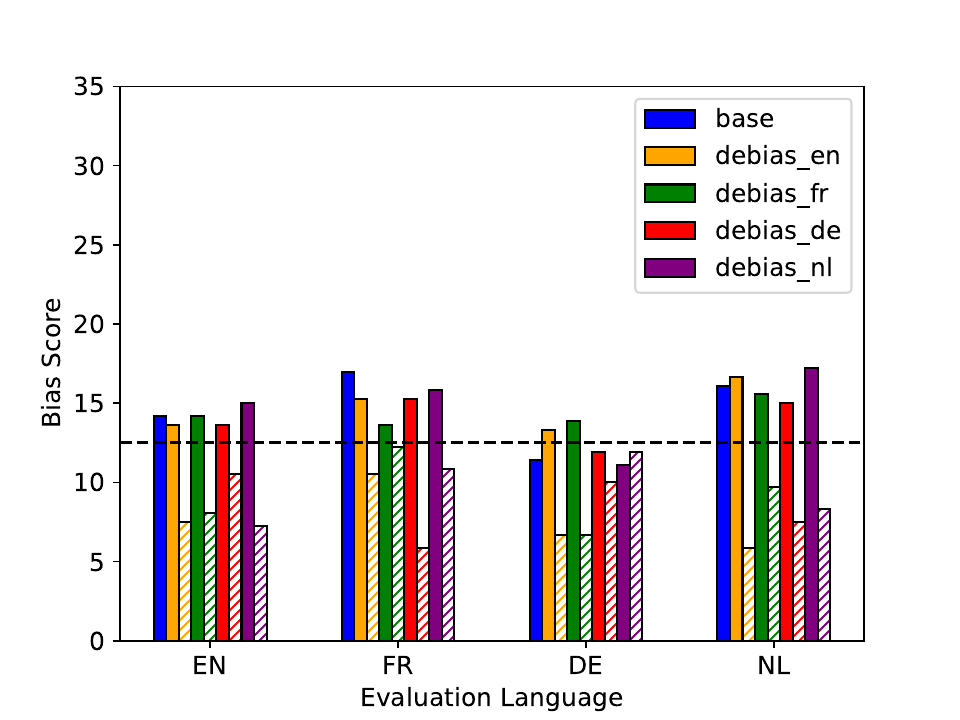}
         \caption{INLP}
     \end{subfigure}
     \hfill
     \begin{subfigure}[t]{0.49\textwidth}
         \centering
         \includegraphics[width=\textwidth]{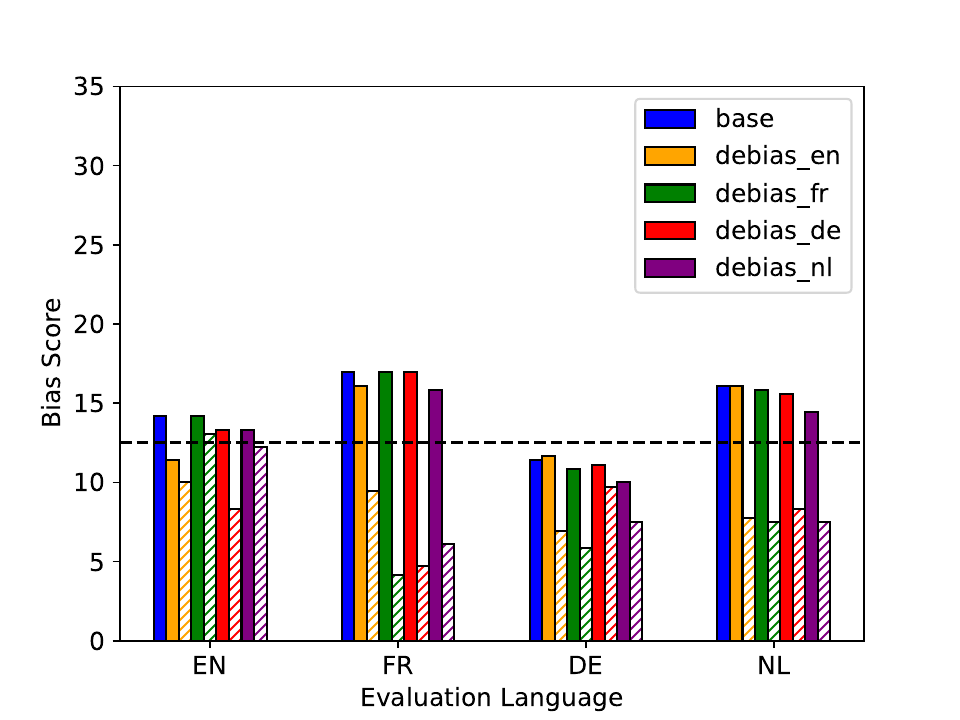}
         \caption{SentDebias} 
     \end{subfigure}
\end{adjustbox}
\caption{Average bias score (across bias types) on different evaluation language when applying INLP and SentDebias in the original space and latent space respectively, with different debiasing languages. The horizontal line at score of 12.5 refers to the significant threshold. Color represents different debiasing languages. Solid bar refers to the result of debiasing in original space and dashed bar refers to debiasing in the latent space. Blue bar refers to the base performance without debiasing. Unbiased model would have bias score of 0.}            
\label{fig:average_aya_exp}
\vspace{-10pt}
\end{figure*}

\begin{table}[h!]
\centering
\resizebox{1\columnwidth}{!}{%
\begin{tabular}{|c|c|cc|cc|}
\hline
          & \multicolumn{1}{c|}{}     & \multicolumn{2}{c|}{INLP}                      & \multicolumn{2}{c|}{SentDebias}               \\ \hline
Eval Lang & \multicolumn{1}{c|}{Base} & \multicolumn{1}{c|}{Original} & Latent         & \multicolumn{1}{c|}{Original} & Latent        \\ \hline
en        & 14.17                     & \multicolumn{1}{c|}{13.61}    & \textbf{7.5}   & \multicolumn{1}{c|}{11.39}    & \textbf{10}   \\ \hline
fr        & 16.94                     & \multicolumn{1}{c|}{15.28}    & \textbf{10.56} & \multicolumn{1}{c|}{16.11}    & \textbf{9.44} \\ \hline
de        & 11.39                     & \multicolumn{1}{c|}{13.33}    & \textbf{6.67}  & \multicolumn{1}{c|}{11.67}    & \textbf{6.94} \\ \hline
nl        & 16.1                      & \multicolumn{1}{c|}{16.67}    & \textbf{5.83}  & \multicolumn{1}{c|}{16.11}    & \textbf{7.78} \\ \hline
\end{tabular}%
}
\caption{Average bias score of Aya-expanse-8B on different evaluation languages when performing INLP/SentDebias in the original space and latent space respectively, with English as debiasing language. The score is averaged across three different bias types.} 
\label{tab:english_debias_exp}
\vspace{-14pt}
\end{table} 


\subsection{Results}
As discussed in previous works \citep{ahn-oh-2021-mitigating, reusens-etal-2023-investigating}, when initial bias scores are already near optimal, further debiasing can cause overcompensation,\footnote{Overcompensation has also been observed in our experiments when the base bias score is not significant. Full details can be found in appendix.} instead amplifying bias. In this regard, we focus on scenarios that bias scores are significant. We utilize the binomial distribution to determine the threshold for significant bias at a 95\% confidence level. For an evaluation set of 40 examples per sample, the threshold is calculated to be 62.5\%. A model exceeding this threshold is considered to exhibit a statistically significant deviation (>12.5) from the unbiased 50\% performance, indicating significant bias\footnote{The calculation process is provided in the appendix.}.

Table \ref{tab:english_debias_exp} shows the performance of different debiasing techniques applied to Aya-expanse with English as the debiasing language in the original space and the learned latent space respectively. The absolute deviation to the ideal unbiased model is reported. The reported bias score is averaged across three bias types (Gender, Religion, Race). Table \ref{tab:english_debias_exp} reveals a slight reduction in bias scores when debiasing with English is applied to the original space. However, this decrease becomes significantly more pronounced when debiasing is performed in the learned latent space. Moreover, performing debiasing within a cross-lingual latent space improves the effectiveness of English-based debiasing when evaluated in other languages, compared to debiasing in the original space. This highlights the improved cross-lingual transferability achieved by leveraging a cross-lingual latent space.

\paragraph{Debiasing with Different Languages} We further investigate the effectiveness of debiasing using other languages (French, German, and Dutch). Figure \ref{fig:average_aya_exp} illustrates the results across different debiasing languages, with a horizontal line at 12.5 indicating the threshold for significant bias. The blue bar represents the base bias score without debiasing. Different colors correspond to different debiasing languages. Solid bars depict the performance of debiasing in the original space, whereas dashed bars represent the results when debiasing is performed in the learned latent space. The figure clearly demonstrates that, no matter the debiasing techniques used, debiasing performed in a cross-lingual latent space leads to substantial improvements compared to debiasing in the original space. This pattern is generally consistent across all tested languages, highlighting both enhanced overall performance and improved cross-lingual transferability. 

As shown in Figure \ref{fig:average_aya_exp}, while English often constitutes a significant portion of the pre-training corpus, it does not always serve as the most effective language for debiasing. Additionally, the language that yields the best debiasing performance is not always the same as the evaluation language. Such patterns hold true for both debiasing in the original vector space and in the latent space. This also aligns with findings in \citet{reusens-etal-2023-investigating}, highlighting the complex cross-lingual interactions within multilingual LLMs.






\section{Conclusion}
We have proposed to perform debiasing in a cross-lingual latent space rather than directly on LLM representations. Our experiments with Aya-expanse, and two debiasing techniques (SentDebias and INLP) across four different languages (English, French, German, and Dutch) have confirmed that this approach is both feasible and effective, particularly when bias scores are significant. Compared to debiasing in the original vector space, our method substantially enhances overall performance across various evaluation languages and improves cross-lingual transferability. 

\section*{Limitations} 
Due to the limited availability of cross-lingual resources for debiasing, our experiments primarily focus on European languages, specifically English, French, German, and Dutch. Expanding this work to non-European languages remains an important direction for future research. Additionally, while we evaluate two widely used debiasing techniques (SentDebias, and INLP), we leave out other methods, such as Dropout Regularization \citep{webster2020measuring}, which could be considered in future studies. Furthermore, although previous research has attempted to minimize noise in attribute word lists, these lists are not exhaustive, making the omission of relevant attributes possible.

\section*{Ethical Consideration}
We do not anticipate any risks in the work. In this study, our use of existing artifacts is consistent with their intended purposes. CrowS-Pairs is licensed under the Creative Commons Attribution-ShareAlike 4.0 International License. 

\section*{Acknowledgement}
We would like to thank all anonymous reviewers for their insightful comments and feedback. This work was supported by DisAI - Improving scientific excellence and creativity in combating disinformation with artificial intelligence and language technologies, a project funded by European Union under the Horizon Europe, GA No. \href{https://doi.org/10.3030/101079164}{101079164}.

\bibliography{acl_latex}

\appendix

\section{Related Work}
\paragraph{Multilingual Learning} Multilingual LLMs have demonstrated impressive capability in solving tasks in different languages, such as text \citep{conneau-etal-2020-unsupervised, xue2020mt5, dang2024aya, dubey2024llama} and code \citep{le2023bloom, chai-etal-2023-ernie, peng-etal-2024-humaneval, nakamura-etal-2025-aurora}. However, they have been observed to exhibit an imperfect cross-lingual alignment, which can be further improved \citep{pan-etal-2021-multilingual, han-etal-2022-x, peng-sogaard-2024-concept}. Such improvements often involve explicit alignment objectives from bilingual dictionary seeds \citep{conneau2017word, chi-etal-2021-improving, li2024prealign}, or by training on mixed corpora constructed using such resources \citep{gouws2015simple, chai2024xcot}. 


\paragraph{Debiasing} The detection and mitigation of bias in LLMs has significant societal impacts. Various types of bias, such as gender, age, and race, have been explored and identified in these models \citep{nangia-etal-2020-crows, wan2023kelly, gallegos-etal-2024-bias}. In the meantime, different debiasing techniques have been proposed to tackle potential biases from different perspectives, including word embeddings \citep{bolukbasi2016man,kaneko-bollegala-2019-gender} and sentence embeddings \citep{liang-etal-2020-towards, kaneko-bollegala-2021-debiasing}, along with prompt-based self-debiasing \citep{10.1162/tacl_a_00434, guo-etal-2022-auto} and model editing \citep{gandikota2024unified}. 

\section{Calculation of the Significant Bias Threshold}
\label{sec:appendix}
In this study, we determine the threshold for significant bias in a binary classification task using statistical hypothesis testing based on the binomial distribution. A perfectly unbiased classifier is expected to achieve a bias score of 50\%, which represents optimal fairness. A higher bias score indicates worse performance, meaning the classifier deviates further from an ideal unbiased state.

To quantify this deviation, we model the number of wrongly classified instances as a binomial random variable, and use normal approximation to calculate the threshold. Given an evaluation set of size $n$, with each instance having an equal probability $p=0.5$ of being classified wrongly under the null hypothesis (unbiased case), the expected number of wrong classifications is:
\begin{equation}
    \mu = np
\end{equation}
with a standard deviation of:
\begin{equation}
    \sigma = \sqrt{np(1-p)}
\end{equation}    

To establish a threshold beyond which a classifier’s bias score is considered significantly worse, we use a one-tailed hypothesis test at a 95\% confidence level ($\alpha=0.05$). The critical threshold $X$ is given by:
\begin{equation}
    X = np + Z_{\alpha} \cdot \sigma
\end{equation}

where $Z_{\alpha}$ is the critical value from the standard normal distribution corresponding to the upper 5\% tail ($Z_{0.05} \approx 1.645$).

For an evaluation set of $n=40$ examples:
\begin{equation}
    X = 40 \times 0.5 + 1.645 \times \sqrt{40 \times 0.5 \times 0.5} \approx 25.2
\end{equation}

Since the number of wrong classifications must be an integer, we round $X$ to 25. Therefore, if it wrongly classifies more than 25 examples (achieving a bias score above 62.5\%), it exhibits a statistically significant deviation from the optimal 50\% performance, and is considered as highly biased. Therefore, the threshold is (62.5\%-50\%) = 12.5\%.

\section{Attribute Lists}
\label{sec:attribute_list}
The different lists of attribute words per language used for SentDebias and INLP are listed below. We use the same attribute lists provided by \citet{reusens-etal-2023-investigating}.

\subsection{English Attribute List}
\paragraph{Gender} actor, actress, actors, actresses, airman, airwoman, airmen, airwomen, uncle, aunt, uncles, aunts, boy, girl, boys, girls, groom, bride, grooms, brides, brother, sister, brothers, sisters, businessman, businesswoman, businessmen, businesswomen, chairman, chairwoman, chairmen, chairwomen, dude, chick, dudes, chicks, dad, mom, dads, moms, daddy, mommy, daddies, mommies, son, daughter, sons, daughters, father, mother, fathers, mothers, male, female, males, females, guy, gal, guys, gals, gentleman, lady, gentlemen, ladies, grandson, granddaughter, grandsons, granddaughters, guy, girl, guys, girls, he, she, himself, herself, him, her, his, her, husband, wife, husbands, wives, king, queen, kings, queens, lord, lady, lords, ladies, sir, ma’am, man, woman, men, women, sir, miss, mr., mrs., mr., ms., policeman, policewoman, prince, princess, princes, princesses, spokesman, spokeswoman, spokesmen, spokeswomen.

\paragraph{Race} black, caucasian, asian, african, caucasian, asian, black, white, asian, africa, america, asia, africa, america, china, africa, europe, asia.

\paragraph{Religion} jewish, christian, muslim, jews, christians, muslims, torah, bible, quran, synagogue, church, mosque, rabbi, priest, imam, judaism, christianity, islam.

\subsection{French Attribute List}
\paragraph{Gender} acteur, actrice, acteurs, actrices, aviateur, aviatrice, aviateurs, aviatrices, oncle, tante, oncle, tantine, tonton, tata, tontons, tatas, oncles, tantes, oncles, tantines, garçon, fille, garçons, filles, marié, mariée, époux, épouse, mariés, mariées, époux, épouses, brother, sister, frère, soeur, frères, soeurs, entrepreneur, entrepreneuse, entrepreneur, entrepreneure, entrepreneurs, entrepreneures, entrepreneurs, entrepreneuses, président, présidente, présidents, présidentes, mec, meuf, gamin, gamine, mecs, meufs, père, mère, pères, mères, papa, maman, papas, mamans, fils, fille, fils, filles, abbé, abbesse, abbés, abbesses, masculin, féminin, mâle, femelle, mâles, femelles, gars, fille, gars, filles, monsieur, dame, messieurs, dames, petit-fils, petite-fille, petit-fils, petites-filles, il, elle, lui-même, elle-même, lui, elle, mari, femme, maris, femmes, roi, reine, rois, reines, seigneur, seigneuresse, seigneurs, seigneuresses, monsieur, m’dame, monsieur, madame, homme, femme, hommes, femmes, monsieur, mademoiselle, mr, mme, mr, mlle, policier, policière, prince,princesse, princes, princesses, copain, copine, copains, copines, ami, amie, amis, amies, voisin, voisine, docteur, doctoresse, docteur, docteure, boulanger, boulangère, héros, héroïne, employé, employée, employés, employées, chef, cheffe, chefs, cheffes, cousin, cousine, grand-pêre, grand-mêre, expert, experte, pompier, pompière, pompiers, pompières, agriculteur, agricultrice, agriculteurs, agricultrices, travailleur, travailleuse, infirmier, infirmière, infirmiers, infirmières, patron, patronne, patrons, patronnes.

\paragraph{Race} noir, blanc, asiatique, black, blanc, asiatique, noir, caucasien, asiatique, africain, européen, asienne, africain, américain, asiatique, afrique, amérique, asie, afrique, amérique, chine, afrique, europe, asie.

\paragraph{Religion} juif, chrétien, musulman, juifs, chrétiens, musulmans, torah, bible, coran, synagogue, église, mosquée, rabbin, prêtre, imam, judaïsme, christianisme, islam.

\subsection{German Attribute List}
\paragraph{Gender} schauspieler, schauspielerin, koch, köchin, lehrer, lehrerin, schüler, schülerin, student, studentin, pilot, pilotin, onkel, tante, junge, mädchen, bräutigam, braut, bruder, schwester, geschäftsmann, geschäftsfrau, vorsitzender, vorsitzende, vater, mutter, papa, mama, sohn, tochter, mann, frau, kerl, mädel, herr, dame, enkel, enkelin, großvater, großmutter, cousin, cousine, er, sie, ihm, ihr, sein, ihr, seine, ihre, ehemann, ehefrau, feuerwehrmann, feuerwehrfrau, könig, königin, fürst, fürstin, herzog, herzogin, mann, frau, männer, frauen, hr., fr., polizist, polizistin, prinz, prinzessin, sprecher, sprecherin, kollege, kollegin, mitarbeiter, mitarbeiterin, helfer, helferin, anwalt, anwältin, bauarbeiter, bauarbeiterin, krankenpfleger, krankenpflegerin, chef, chefin, vorgesetzter, vorgesetzte, sänger, sängerin, kunde, kundin, besucher, besucherin, freund, freundin, arzt, ärztin, verkäufer, verkäuferin, kanzler, kanzlerin, geschäftsleiter, geschäftsleiterin, pfleger, pflegerin, kellner, kellnerin.

\paragraph{Race} dunkelhäutig, hellhäutig, asiatisch, afrikaner, europäer, asiate, amerikaner, afrika, amerika, asien, china.

\paragraph{Religion} jüdisch, christlich, muslimisch, jude, christ, muslim, torah, bibel, koran, synagoge, kirche, moschee, rabbiner, pfarrer, imam, judentum, christentum, islam.

\subsection{Dutch Attribute List}
\paragraph{Gender} acteur, actrice, acteurs, actrices, oom, tante, ooms, tantes, nonkel, tante, nonkels, tantes, jongen, meisje, jongens, meisjes, bruidegom, bruid, bruidegommen, bruiden, broer, zus, broers, zussen, zakenman, zakenvrouw, zakenmannen, zakenvrouwen, kerel, griet, kerels, grieten, vader, moeder, vaders, moeders, papa, mama, papa’s, mama’s, zoon, dochter, zonen, dochters, man, vrouw, mannen, vrouwen, gast, wijf, gasten, wijven, heer, dame, heren, dames, kleinzoon, kleindochter, kleinzonen, kleindochters, vent, vrouw, venten, vrouwen, hij, zij, hemzelf, haarzelf, hem, haar, zijn, haar, mannelijk, vrouwelijk, vriend, vriendin, vrienden, vriendinnen, koning, koningin, koningen, koninginnen, heer, dame, heren, dames, meneer, mevrouw, jongeheer, jongedame, jongeheren, jongedames, jongeheer, juffrouw, jongeheren, juffrouwen, politieagent, politieagente, prins, prinses, prinsen, prinsessen, woordvoerder, woordvoerster, woordvoerders, woordvoersters, brandweerman, brandweervrouw, brandweermannen, brandweervrouwen, timmerman, timmervrouw, timmermannen, timmervrouwen, meester, juf, meesters, juffen, verpleger, verpleegster, verplegers, verpleegsters, bestuurder, bestuurster, bestuurders, bestuursters, kuisman, kuisvrouw, kuismannen, kuisvrouwen, kok, kokkin, kokken, kokkinnen, leraar, lerares, directeur, directrice, directeurs, directrices, secretaris, secretaresse, secretarissen, secretaressen, boer, boerin, boeren, boerinnen, held, heldin, gastheer, gastvrouw, gastheren, gastvrouwen, opa, oma, opa’s, oma’s, grootvader, grootmoeder, grootvaders, grootmoeders.

\paragraph{Race} afrikaans, amerikaans, aziatisch, afrikaans, europees, aziatisch, zwart, blank, aziatisch, afrika, amerika, azië, afrika, amerika, china, afrika, europa, azië.

\paragraph{Religion} joods, christen, moslim, joden, christenen, moslims, thora, bijbel, koran, synagoge, kerk, moskee, rabbijn, priester, imam, jodendom, christendom, islam.

\section{Autoencoder Architecture}
\begin{figure*}[h!]
     \centering
     \begin{adjustbox}{minipage=\columnwidth,scale=1} 
     \begin{subfigure}[t]{0.49\columnwidth}
         \centering
         \includegraphics[width=\columnwidth]{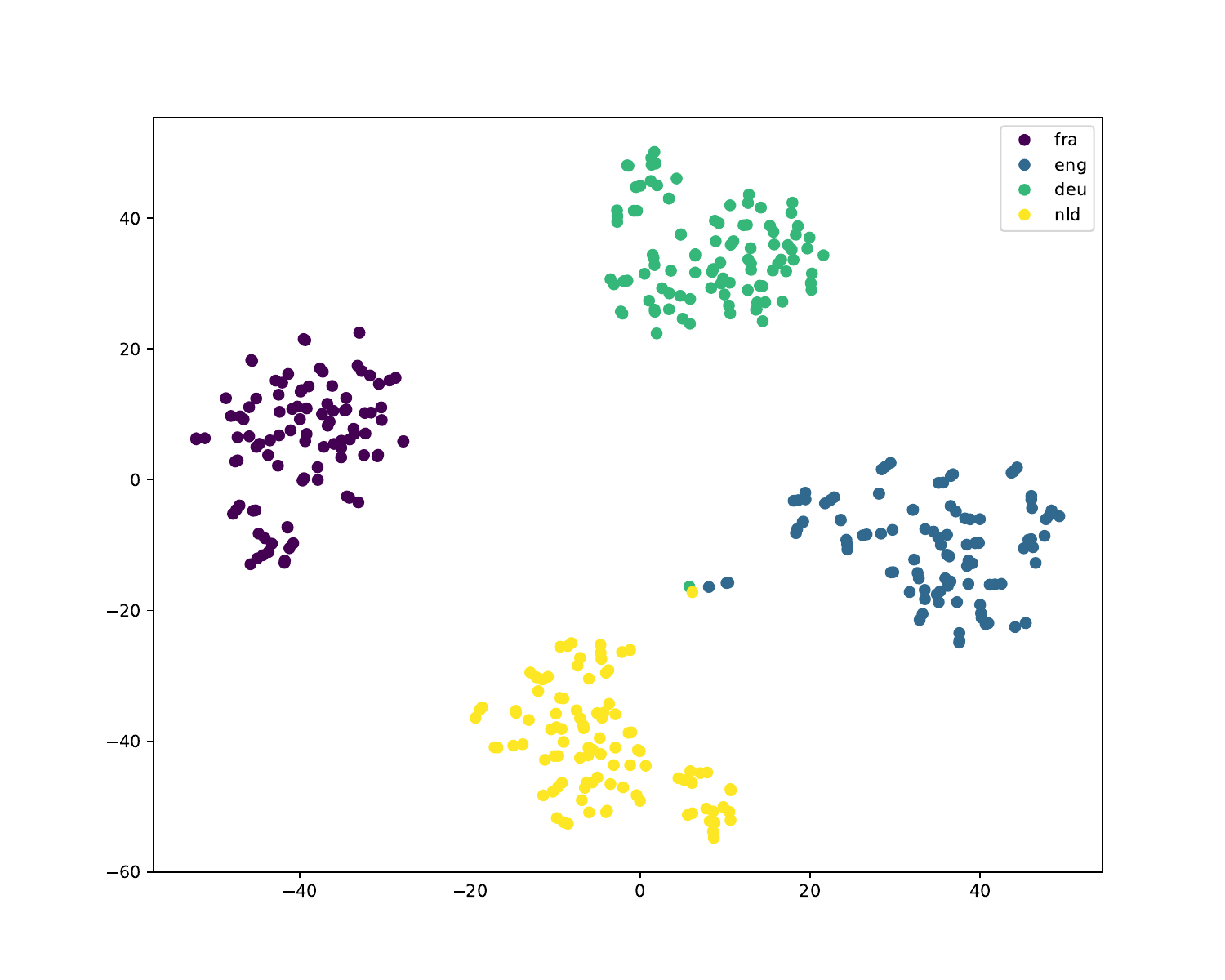}
         \caption{Original Space}
    \end{subfigure}
    \hfill
    \begin{subfigure}[t]{0.49\columnwidth}
         \centering
         \includegraphics[width=\columnwidth]{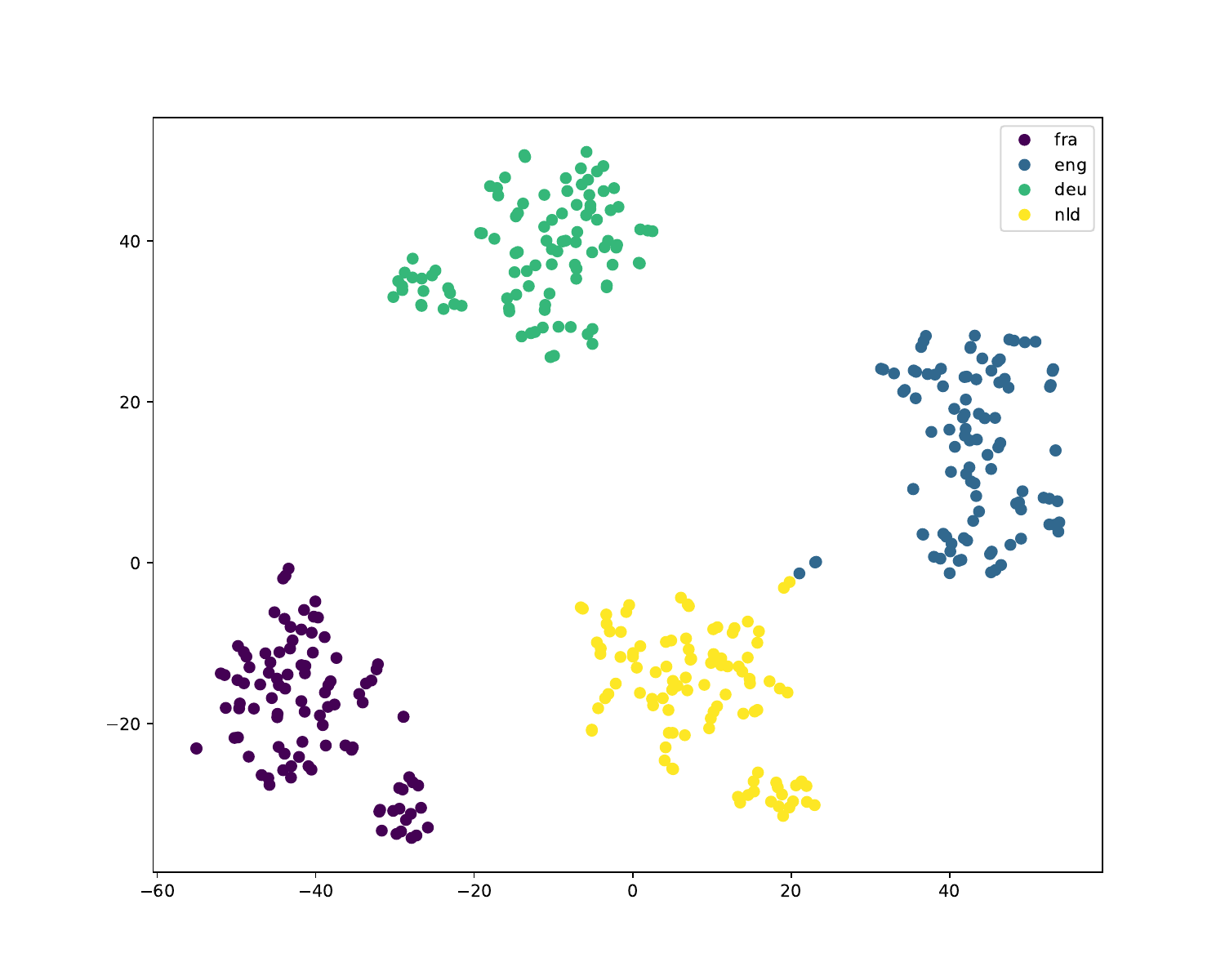}
         \caption{AE with 1-layer MLP}
    \end{subfigure}
    \newline
    \begin{subfigure}[t]{0.49\columnwidth}
         \centering
         \includegraphics[width=\columnwidth]{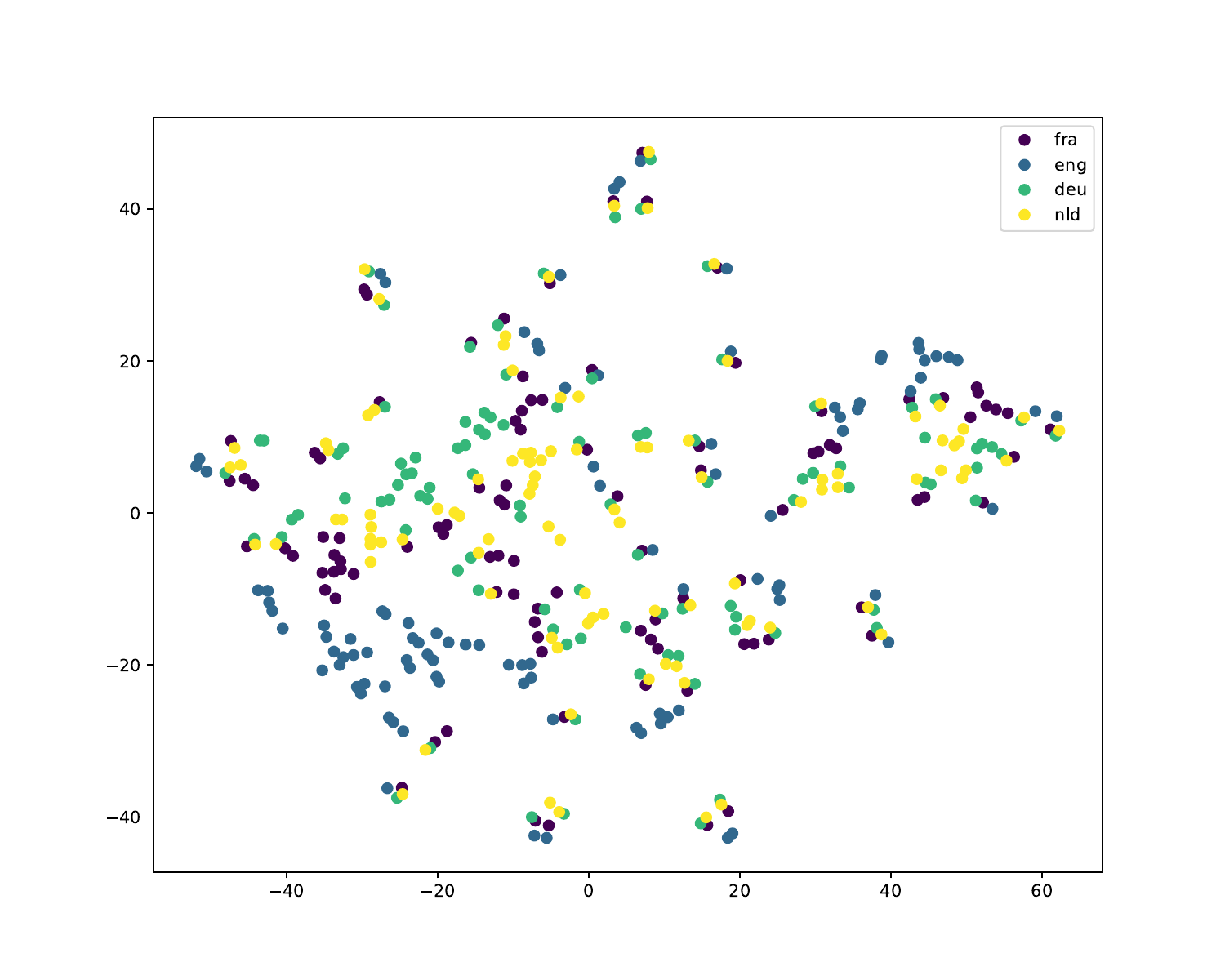}
         \caption{AE with 2-layer MLP}
    \end{subfigure}
    \hfill
    \begin{subfigure}[t]{0.49\columnwidth}
         \centering
         \includegraphics[width=\columnwidth]{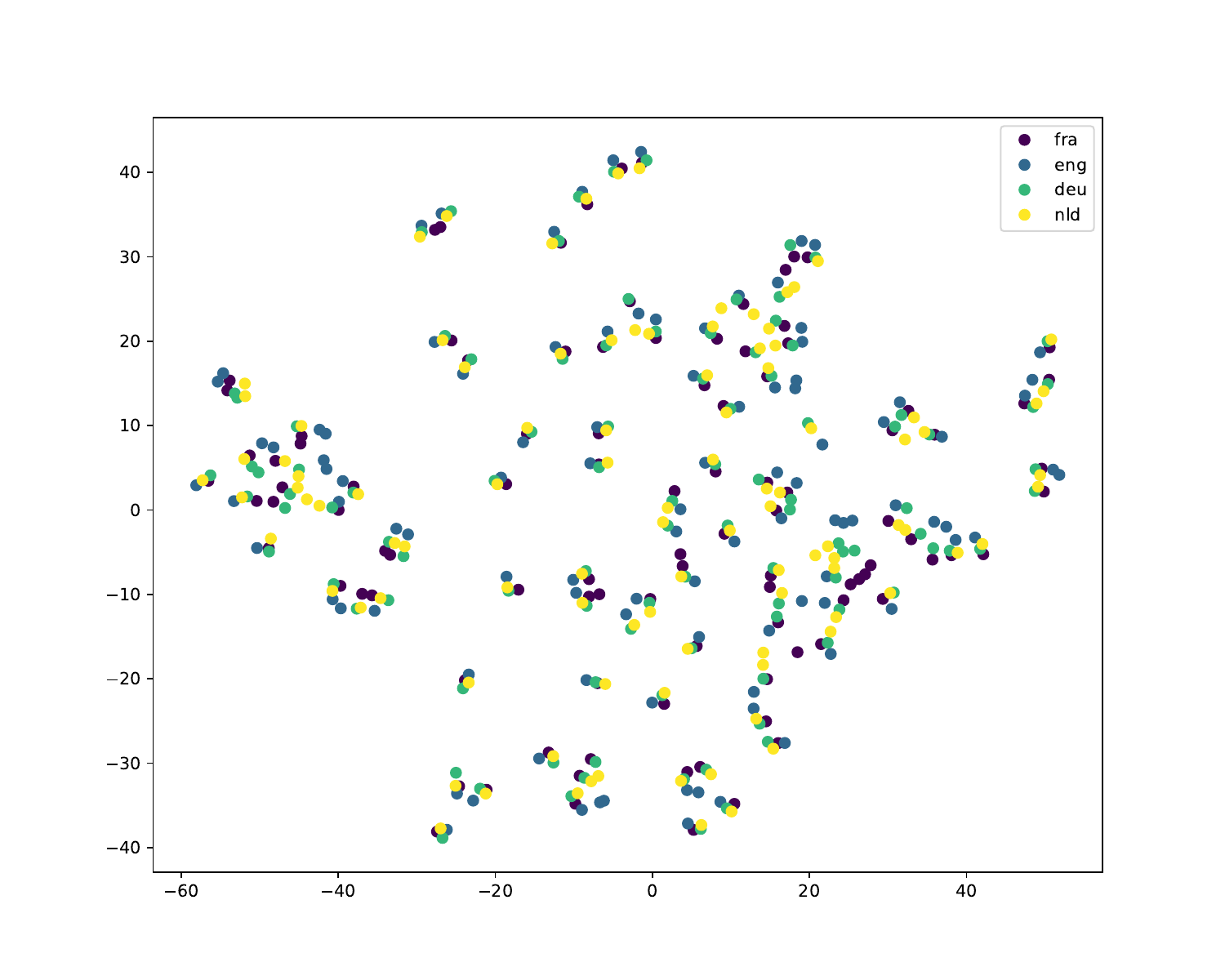}
         \caption{AE with 3-layer MLP}
    \end{subfigure}
    \newline
    \hspace*{.25\columnwidth}
    \begin{subfigure}[t]{0.49\columnwidth}
         \centering
         \includegraphics[width=\columnwidth]{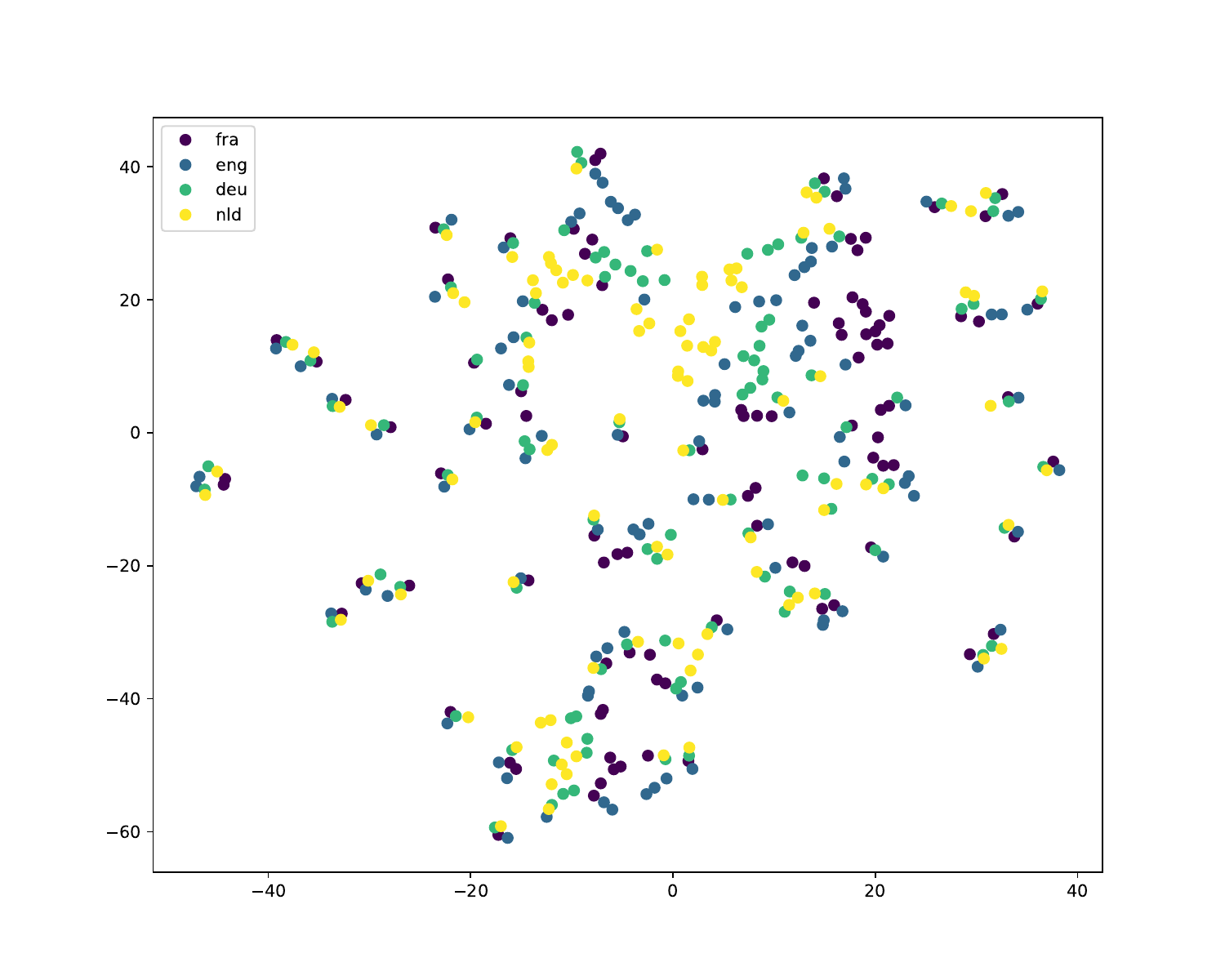}
         \caption{AE with 4-layer MLP}
    \end{subfigure}

\end{adjustbox}
\caption{The t-SNE visualization of 100 parallel sentences across four languages taken from Flores+ in the original space and latent spaces induced by autoencoder with different hidden layers respectively.}
\label{fig:ae_architecture}
\end{figure*}

We have experimented with autoencoders that have one, two, three, and four layers. The visualization of 100 parallel sentences can be found in Figure \ref{fig:ae_architecture}. It clearly shows that the autoencoder with 3-layer MLP gives the best cross-lingual alignment. So we choose autoencoder with 3-layer MLP for our experiments. To perform debiasing in the latent space, we first use encoders to encode the representation into the cross-lingually aligned latent space. The debiasing technique is then performed in it. After debiasing, the representation is projected back using the decoder for further generation. 

\section{Full Experimental Results}
\begin{figure*}[h!]
\vspace{-5pt}
     \centering
     \begin{adjustbox}{minipage=\textwidth,scale=0.8} 
    \begin{subfigure}[t]{0.3\textwidth}
         \centering
         \includegraphics[width=\textwidth]{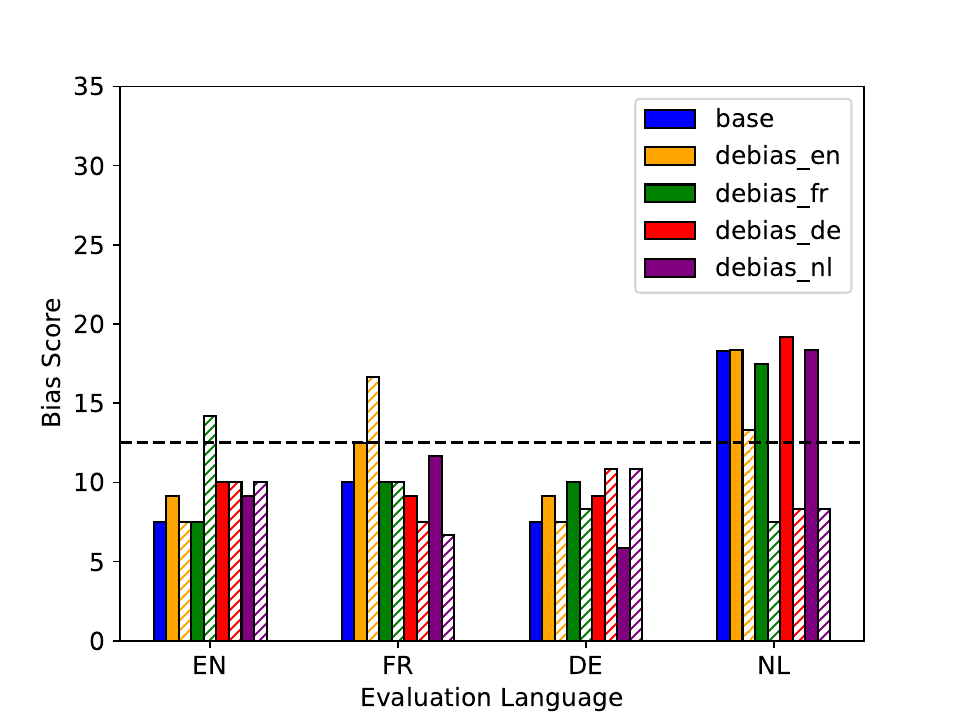}
         \caption{INLP (Race)}
     \end{subfigure}
     \hfill
     \begin{subfigure}[t]{0.3\textwidth}
         \centering
         \includegraphics[width=\textwidth]{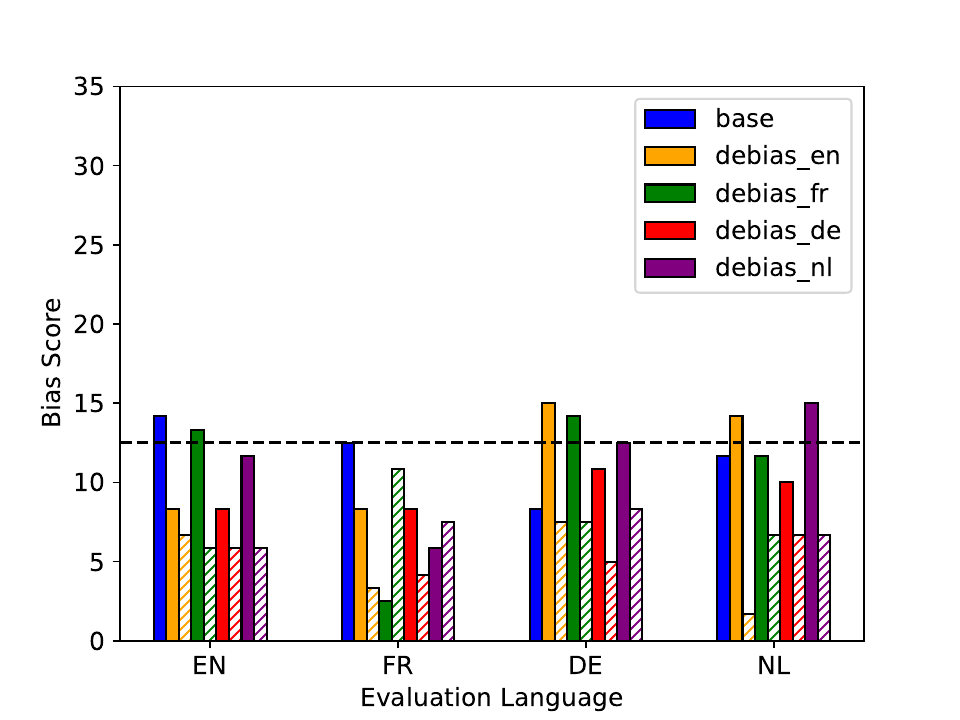}
         \caption{INLP (Gender)} 
     \end{subfigure}
    \hfill
    \begin{subfigure}[t]{0.3\textwidth}
         \centering
         \includegraphics[width=\textwidth]{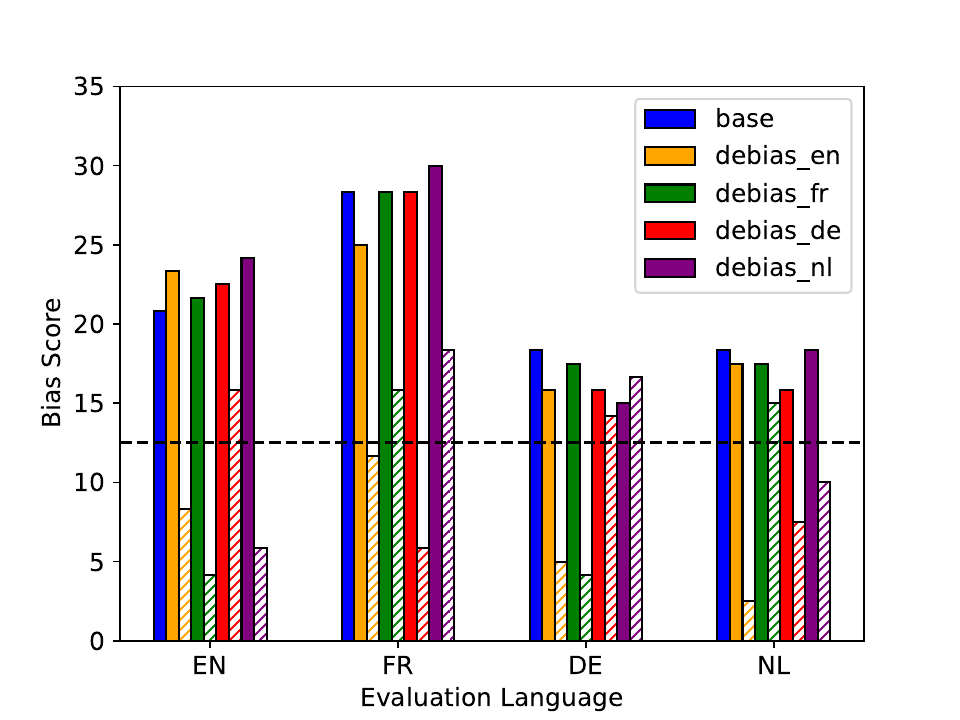}
         \caption{INLP (Religion)}
    \end{subfigure}
    \newline
     \begin{subfigure}[t]{0.3\textwidth}
         \centering
         \includegraphics[width=\textwidth]{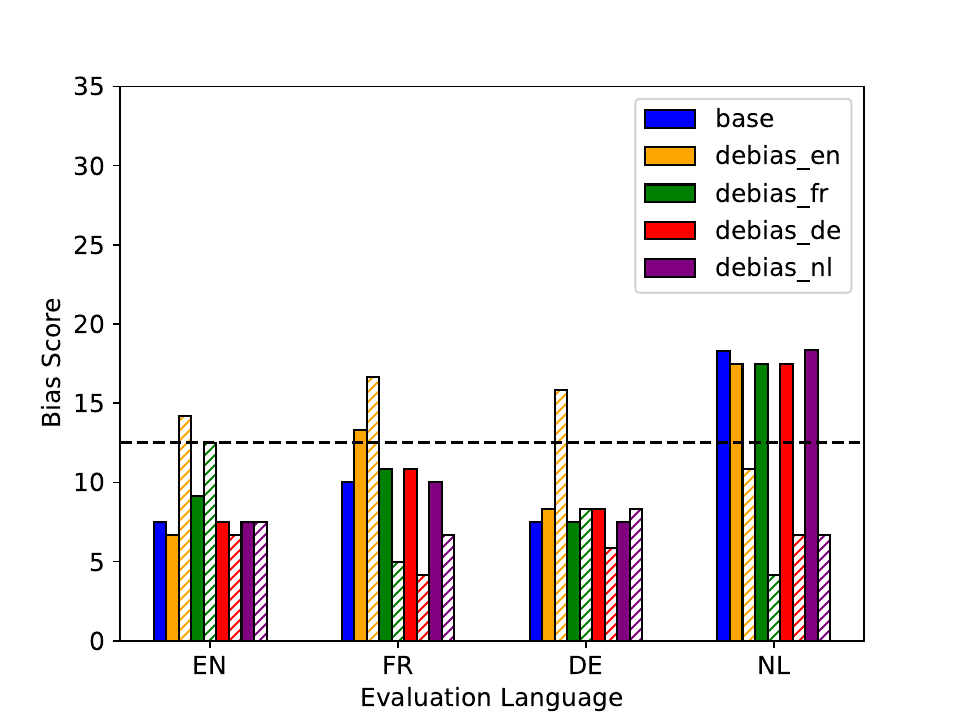}
         \caption{SentDebias (Race)}
     \end{subfigure}
     \hfill
     \begin{subfigure}[t]{0.3\textwidth}
         \centering
         \includegraphics[width=\textwidth]{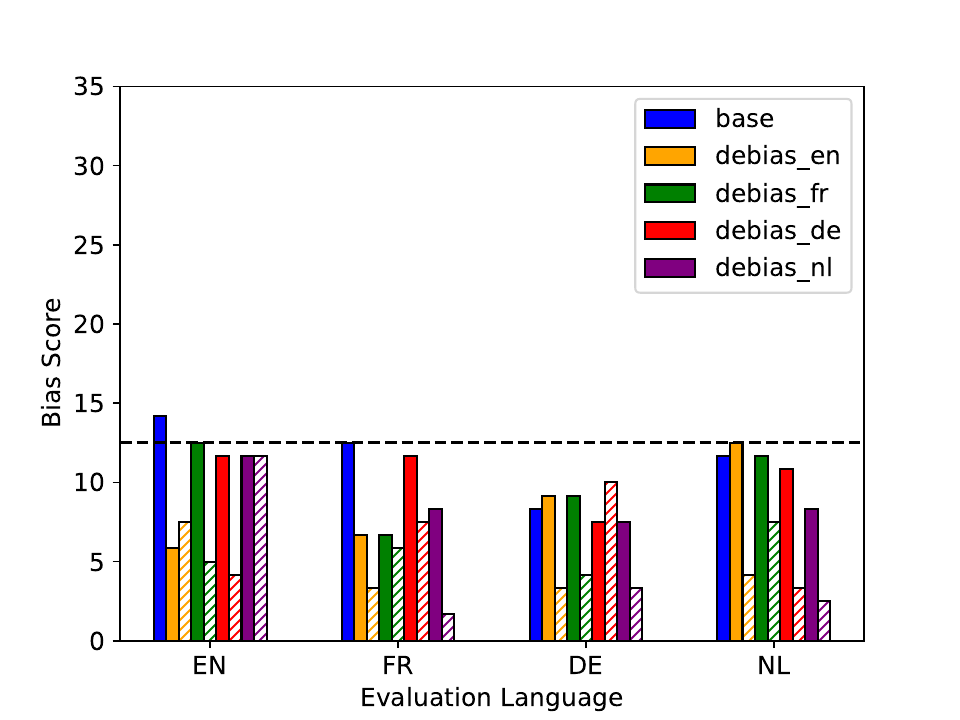}
         \caption{SentDebias (Gender)} 
     \end{subfigure}
    \hfill
    \begin{subfigure}[t]{0.3\textwidth}
         \centering
         \includegraphics[width=\textwidth]{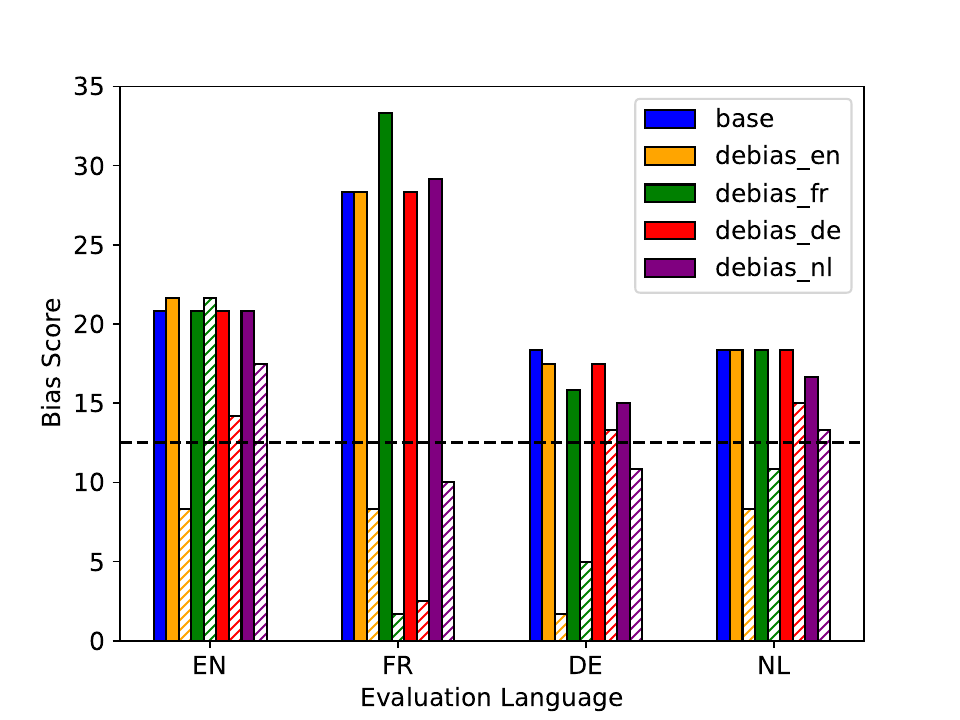}
         \caption{SentDebias (Religion)}
    \end{subfigure}
\end{adjustbox}
\caption{Bias score of Aya-expanse-8B on different evaluation language when applying INLP/SentDebias in the original space and latent space respectively, with different debiasing languages. The horizontal line at score of 12.5 refers to the significant threshold. Color represents different debiasing languages. Solid bar refers to the result of debiasing in original space and dashed bar refers to debiasing in the latent space. Blue bar refers to the base performance without debiasing.}         
\label{fig:aya_exp}
\end{figure*}

\begin{figure*}[h!]
\vspace{-5pt}
     \centering
     \begin{adjustbox}{minipage=\textwidth,scale=0.9} 
     \begin{subfigure}[t]{0.3\textwidth}
         \centering
         \includegraphics[width=\textwidth]{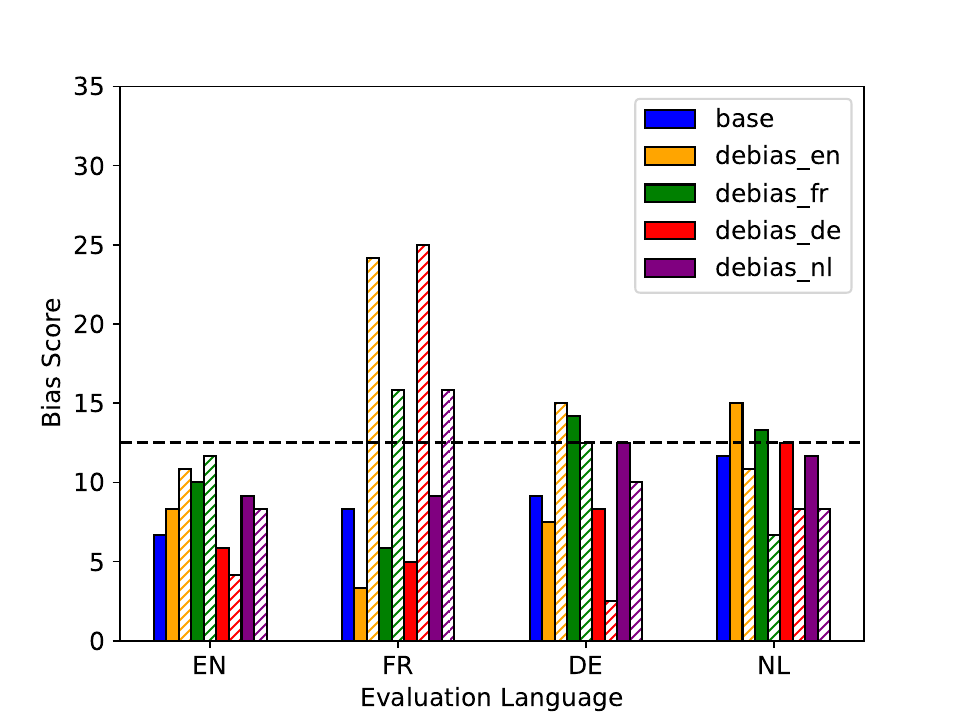}
         \caption{INLP (Race)}
    \end{subfigure}
    \hfill
    \begin{subfigure}[t]{0.3\textwidth}
         \centering
         \includegraphics[width=\textwidth]{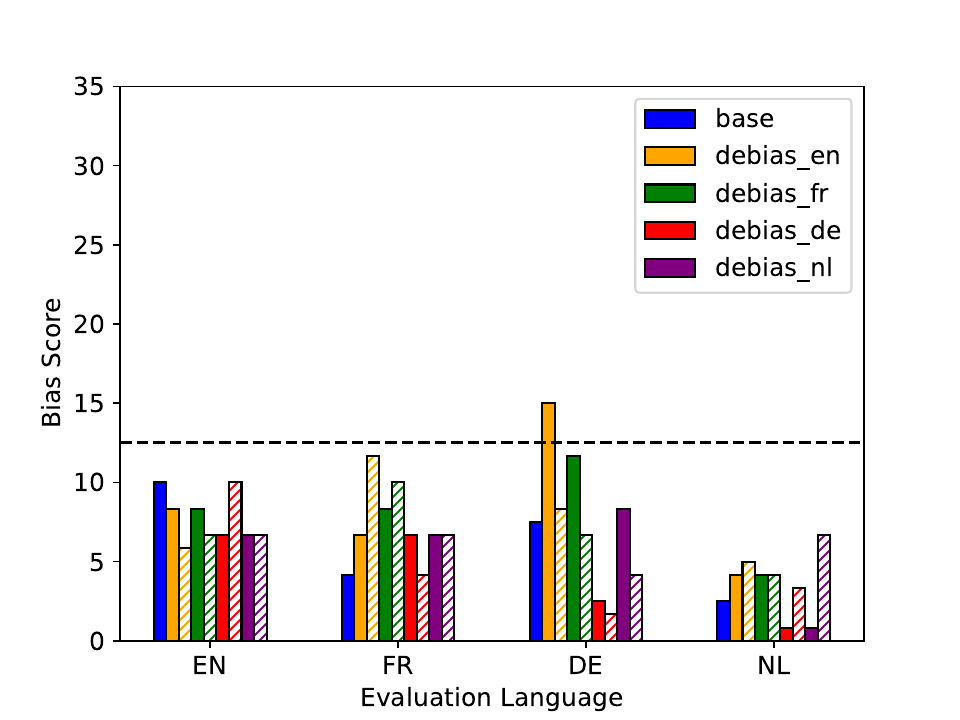}
         \caption{INLP (Gender)}
    \end{subfigure}
    \hfill
    \begin{subfigure}[t]{0.3\textwidth}
         \centering
         \includegraphics[width=\textwidth]{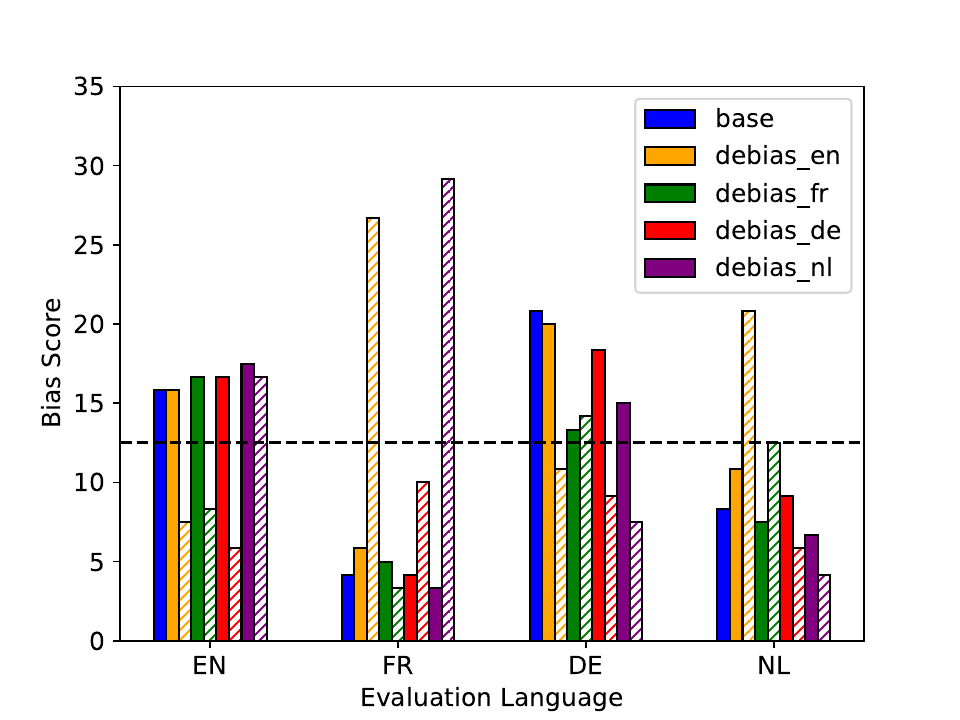}
         \caption{INLP (Religion)}
    \end{subfigure}
    \newline
    \begin{subfigure}[t]{0.3\textwidth}
         \centering
         \includegraphics[width=\textwidth]{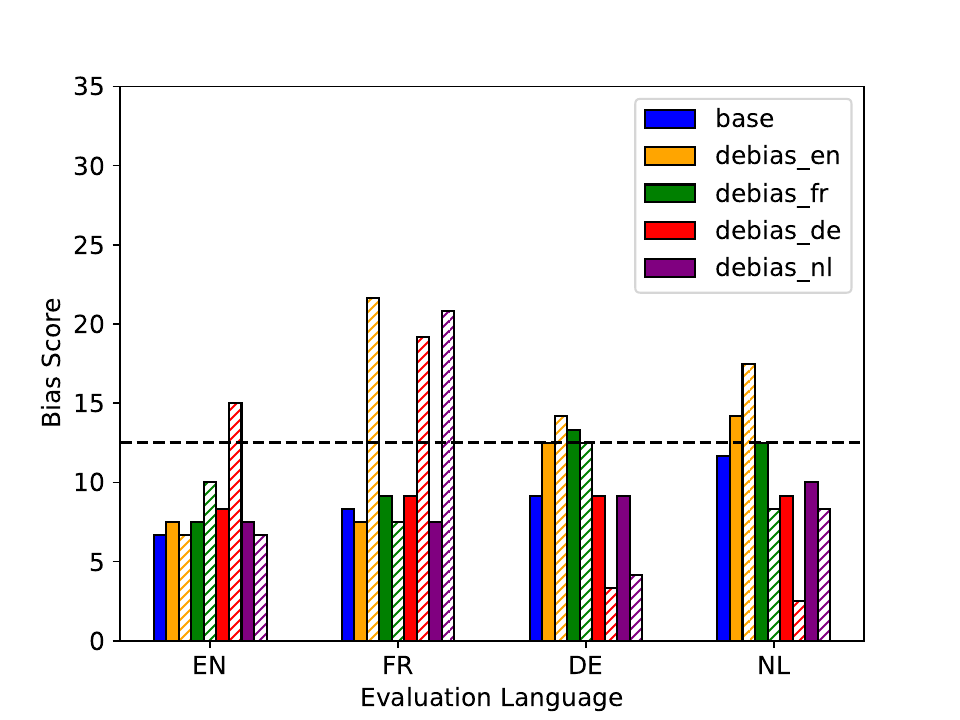}
         \caption{SentDebias (Race)}
     \end{subfigure}
     \hfill
     \begin{subfigure}[t]{0.3\textwidth}
         \centering
         \includegraphics[width=\textwidth]{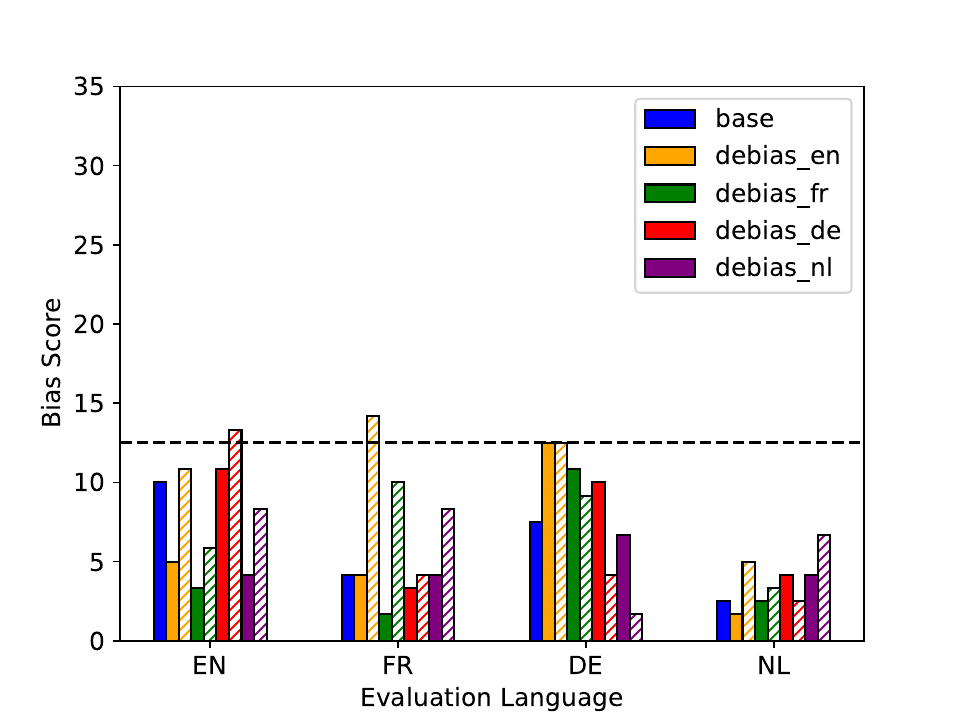}
         \caption{SentDebias (Gender)} 
     \end{subfigure}
    \hfill
    \begin{subfigure}[t]{0.3\textwidth}
         \centering
         \includegraphics[width=\textwidth]{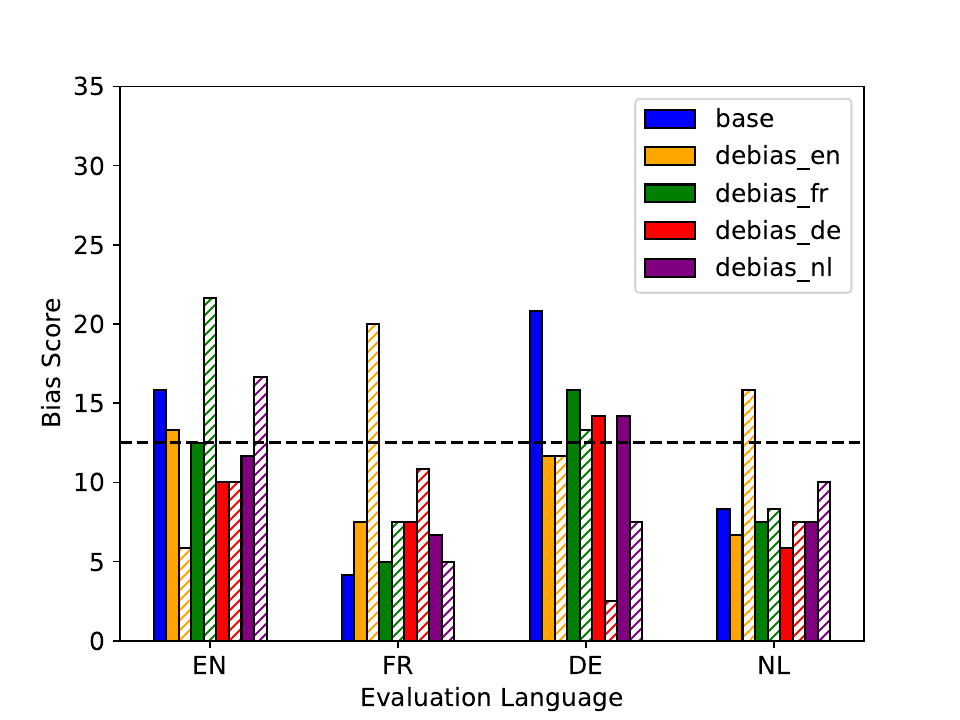}
         \caption{SentDebias (Religion)}
    \end{subfigure}
\end{adjustbox}
\caption{Bias score of Llama3.1-8B-Instruct on different evaluation language when performing INLP/SentDebias in the original space and latent space respectively, with different debiasing languages. The horizontal line at score of 12.5 refers to the significant threshold. Color represents different debiasing languages. Solid bar refers to the result of debiasing in original space and dashed bar refers to debiasing in the latent space. Blue bar refers to the base performance without debiasing.}   
\label{fig:llama_exp}
\end{figure*}

In the Appendix, we report our full experimental results across different types of bias in Figure \ref{fig:aya_exp} and \ref{fig:llama_exp}. It is noticeable that the debiasing is effective when the bias is significant. All experiments are run on a single NVIDIA A100 GPU.

\section{Impact on Downstream Evaluation}
As already discussed in \citet{meade-etal-2022-empirical}, most of the models are relatively unaffected by debiasing (with debiasing techniques like SentDebias and INLP). Similarly, our preliminary experiments on GLUE (SST-2, RTE, and MRPC) \citep{wang-etal-2018-glue} indicate that debiasing in the latent space does not significantly degrade model performance. The relative performance changes are minimal: only a 1.3\% difference in F1 score for MRPC, 0.2\% in accuracy for SST-2, and 0.77\% in accuracy for RTE. These results are comparable to the effects observed when debiasing is applied in the original representation space.

\end{document}